\documentclass[10pt,twocolumn,letterpaper]{article}

\usepackage{cvpr23/cvpr}              % To produce the CAMERA-READY version
\usepackage{graphicx}
\usepackage{subcaption}
\usepackage{amsmath,amsfonts,bm}
\usepackage{amssymb}
\usepackage{booktabs}
\usepackage{makecell}
\usepackage{multirow}
\usepackage{xcolor}
\usepackage{ulem}

\usepackage[pagebackref,breaklinks,colorlinks]{hyperref}

\usepackage[capitalize]{cleveref}
\crefname{section}{Sec.}{Secs.}
\Crefname{section}{Section}{Sections}
\Crefname{table}{Table}{Tables}
\crefname{table}{Tab.}{Tabs.}

\begin{document}

%%%%%%%%% TITLE - PLEASE UPDATE
\title{VDN-NeRF: Resolving Shape-Radiance Ambiguity via View-Dependence Normalization}

\renewcommand{\thefootnote}{\fnsymbol{footnote}}

\author{
Bingfan Zhu$^{\ast1}$ \quad 
Yanchao Yang$^{\ast\dagger2,3}$ \quad 
Xulong Wang$^1$ \quad 
Youyi Zheng$^{\dagger1}$ \quad 
Leonidas Guibas$^3$ \\
$^1$Zhejiang University\footnotemark[3] \quad $^2$The University of Hong Kong\footnotemark[4] \quad $^3$Stanford University
}
% \author{First Author\\
% Institution1\\
% Institution1 address\\
% {\tt\small firstauthor@i1.org}
% % For a paper whose authors are all at the same institution,
% % omit the following lines up until the closing ``}''.
% % Additional authors and addresses can be added with ``\and'',
% % just like the second author.
% % To save space, use either the email address or home page, not both
% \and
% Second Author\\
% Institution2\\
% First line of institution2 address\\
% {\tt\small secondauthor@i2.org}
% }
\maketitle

\footnotetext[1]{Equal contributions.}
\footnotetext[2]{Corresponding authors (yanchaoy@hku.hk, youyizheng@zju.edu.cn).}
\footnotetext[3]{The State Key Lab of CAD\&CG.}
\footnotetext[4]{The department of EEE and the Institute of Data Science.}

\renewcommand{\thefootnote}{\arabic{footnote}}

%%%%%%%%% ABSTRACT
\begin{abstract}
We propose VDN-NeRF, a method to train neural radiance fields (NeRFs) for better geometry under non-Lambertian surface and dynamic lighting conditions that cause significant variation in the radiance of a point when viewed from different angles. Instead of explicitly modeling the underlying factors that result in the view-dependent phenomenon, which could be complex yet not inclusive, we develop a simple and effective technique that normalizes the view-dependence by distilling invariant information already encoded in the learned NeRFs. We then jointly train NeRFs for view synthesis with view-dependence normalization to attain quality geometry. Our experiments show that even though shape-radiance ambiguity is inevitable, the proposed normalization can minimize its effect on geometry, which essentially aligns the optimal capacity needed for explaining view-dependent variations. Our method applies to various baselines and significantly improves geometry without changing the volume rendering pipeline, even if the data is captured under a moving light source. Code is available at: \hyperref[https://github.com/BoifZ/VDN-NeRF]{\color{magenta}{https://github.com/BoifZ/VDN-NeRF}}. 
\end{abstract}

%%%%%%%%% BODY TEXT
\section{Introduction}
\label{sec:intro}

\begin{figure}
    \centering
    \includegraphics[width=\linewidth]{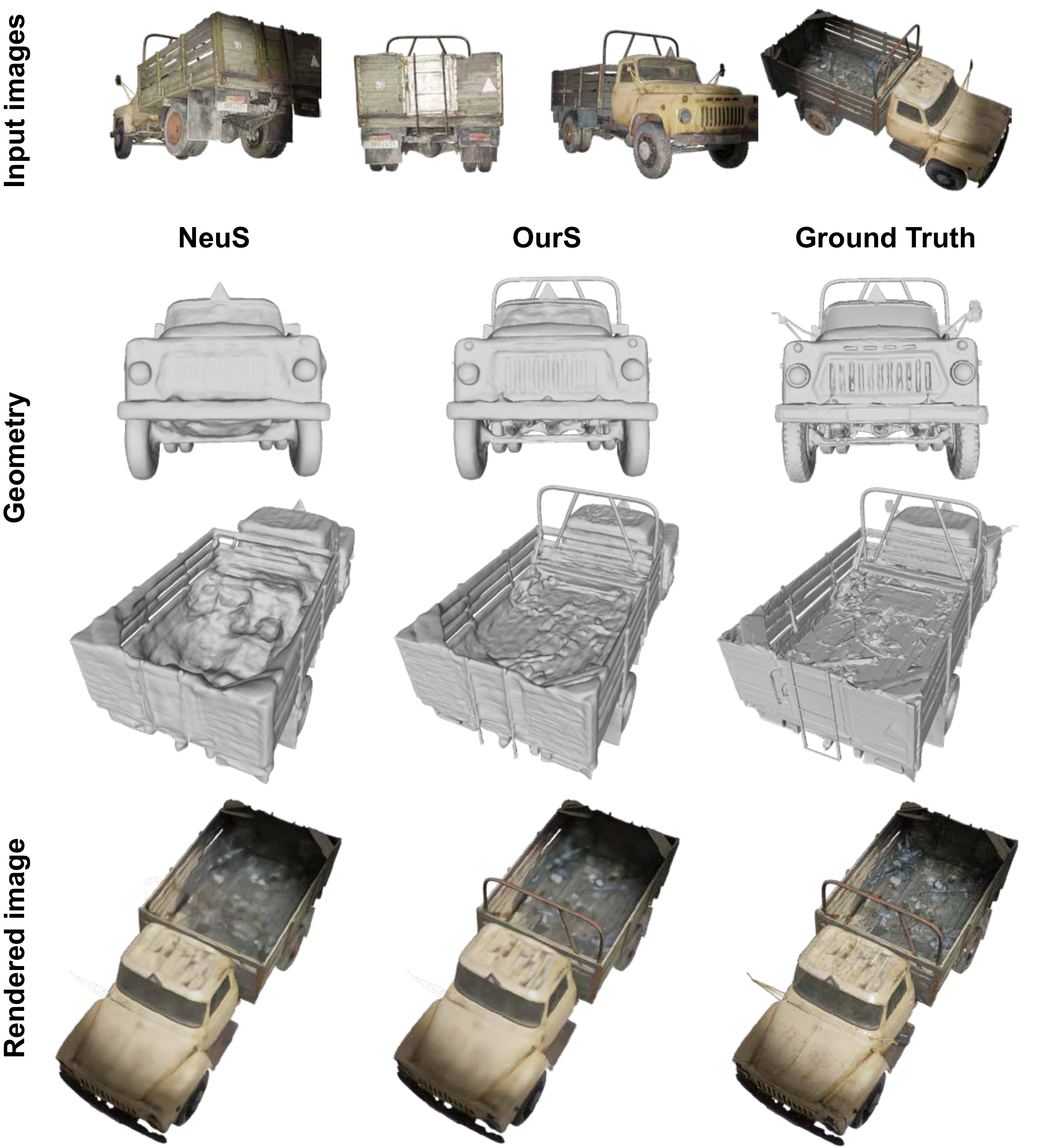}
    \label{fig:teaser}
    \caption{We aim for quality geometry with NeRF reconstruction under inconsistency when viewing the same 3D point, for example, with images captured under a dynamic light field (top row), shown by the light spots on the truck cast from a torch moving with the camera. The middle two rows compare the reconstructed geometry from NeuS \cite{wang2021neus} (first column) and our method (second column). As observed, our method produces more details and better estimates the truck's structure. The last row shows a novel view rendering from both methods. Even though our method normalizes the view-dependence, it does not lose details on the synthesized images thanks to the regularity induced by better geometry.}
\end{figure}

Reconstructing the geometry and appearance of a 3D scene from a set of 2D images is one of the fundamental tasks in computer vision and graphics. 
Recently, based on volume rendering, neural radiance fields (NeRFs) have shown great potential in capturing detailed appearance of the scene, evidenced by their high-quality view synthesis. However, the reconstructed geometry is still far from satisfying. 
Since geometry is critical for physical interaction leveraging such scene representations, many recent works have started to improve the geometric reconstruction within the context of volume rendering,
e.g., with surface rendering or external signals that are more informative of the geometry, like depth scans.

Nevertheless, most improvements do not explicitly study the shape-radiance ambiguity that could induce degenerated geometric reconstructions.
This ambiguity persists as long as some capacity is needed to account for the directional variations in the radiance, which is further amplified if the ambient light field changes according to the observer's viewpoint.
For example, the NeRF's Multi-Layer Perceptrons (MLPs) takes in a  3D location and a 2D direction vector and outputs the observed color of this point from this specific viewing angle.
If the MLP has sufficient capacity to model the directional phenomenon, a perfect photometric reconstruction could be achieved even if the learned geometry is entirely wrong. 
In other words, wrong geometry incurs more directional variations, which the MLP's view-dependent branch can still encode, thus, rendering the photometric reconstruction loss incapable of constraining the solutions.

On the other hand, directional capacity is needed to prevent the (view-dependent) photometric loss from distorting the geometry (confirmed through our experiments).
With quality geometry as the central goal, 
we are now facing a tradeoff between the capacity of the view-dependent radiance function and the shape-radiance ambiguity.
Namely,  the MLP's capacity has to be increased to explain the directional variations, but if the capacity gets too large, shape-radiance ambiguity will permit degenerated geometric reconstructions.
One can thus tune the capacity to achieve the best geometry for each scene, but it is time-consuming and infeasible since different scenes come with different levels of view dependence.

Instead of tuning the directional capacity, we adjust the view-dependence. More explicitly, we propose to normalize the directional variations by encoding invariant features distilled from the same scene representation. 
By doing so, the view-dependence of different scenes is aligned to the same level so that a single optimal capacity can ensure good view synthesis and prevent shape-radiance ambiguity from degrading the geometric estimation. 
Given the self-distillation property, the proposed view-dependence normalization can be jointly trained with the commonly used photometric reconstruction loss. It can also be easily plugged into any method relying on volume rendering for geometric scene reconstruction.

We demonstrate the effectiveness of the proposed view-dependence normalization on several datasets. 
Especially we verify that the level of view-dependence affects the minimality of the shape-radiance ambiguity with tunable directional capacity. 
And we show that our method effectively aligns the optimal capacity for each scene.
To evaluate how our method works under a dynamically changing light field, we propose a new benchmark where a light source is moving conditioned on the camera pose.
We validate that the geometry obtained from our method degrades gracefully as the significance of the induced directional variations increases.
In summary, we make the following contributions:
\begin{itemize}
    \item We perform a detailed study of the coupling between the capacity of the view-dependent radiance function and the shape-radiance ambiguity in the context of volume rendering.
    \item We propose a view-dependence normalization method that effectively aligns the optimality of the directional function capacity under shape-radiance ambiguity for each scene and achieves state-of-the-art geometry compared to various baselines.
    \item We quantitatively verify the robustness of our method when the light field changes to further increase the directional variation, ensuring the applicability to scenarios with unfavored lighting conditions.
\end{itemize}

% % IDR
% Aiming at high-quality surface extraction, others \zbf{[][]} models the surface with the zero-level set of a signed distance function (SDF) 
% recent works [33, 39] use signed distance functions (SDF) [25] for surface representation and introduce the SDF-induced density function to enable the volume rendering to learn an implicit SDF representation.

% \yanchao{this reads like related work, can be moved to related work section}\cite{oechsle2021unisurf,wang2021neus} combines the advantage of surface rendering based and volume rendering based methods, enabling the reconstruction of accurate geometry from multi-view images without masks. NeuS[] further proposed a well-designed weight function which is unbiased and occlusion-aware.
% [Geo-NeuS] focuses on geometry consistent surfaces reconstruction, extend NeuS with multi-view geometry constraints so as to obtain.

% \zbf{[][]} extends neural radiance fields to a dynamic domain. With time as additional input, D-NeRF \cite{pumarola_2021_dnerf} brought the benefits into non-rigid and time-varying scenes modeling.

% In order to render novel appearance for objects, \zbf{[]\cite{oechsle_2020_lightfields}} take advantage of variational autoencoder (VAE) to learn an implicit representation of conditional implicit surface light fields.

\section{Related Work}
\label{sec:related}

Leveraging volume rendering, NeRF \cite{mildenhall_2020_nerf} represents a scene as a continuous volumetric function parameterized by MLPs, which explicitly models volume density and view-dependent radiance computed from a 5D coordinate. It enables promising applications with photo-realistic rendering from arbitrary views.
Follow-up works \cite{zhang2020nerf++,wang_2021_nerfmm,barron_2021_mipnerf,yu2021pixelnerf,lin2021barf,jeong2021scnerf,jain2021putting,niemeyer2022regnerf} improve NeRF to achieve high-quality novel view rendering, however, the quality of the reconstructed geometry still lags.
%the learnable volumetric density function facilitates surface reconstruction through level sets.
%, they are always noisy and lack fidelity.
%This may due to absence of global constraint. 
%Note alpha values are always computed by normalization on a single sampling ray, so that density is actually a relative value in local area.

% Substantial progress has been made to improve the quality and capabilities of NeRF in novel view synthesis (NVS) \cite{zhang2020nerf++, barron_2021_mip, niemeyer2022regnerf}, Relighting \cite{martin2021nerfw, verbin2022refnerf}, and scene editing \cite{kobayashi2022decomposing, verbin2022refnerf, niemeyer2021giraffe}.

%\textbf{High quality neural surface.}
Many methods thus perform disentanglement of shape and radiance in volume rendering for better appearance and geometry reconstruction. 
UNISURF \cite{oechsle2021unisurf} treats the surface as a decision boundary of a binary occupancy classifier. 
Several approaches \cite{yariv2020idr,yariv2021volume,wang2021neus,darmon2022improving,fu2022geo,jiang2020sdfdiff} employ signed distance functions (SDFs) to better model the shape-induced volumetric rendering weights. 
It is noted in \cite{yariv2020idr} that a global shape condition could be helpful to synthesize the appearance accurately. 
%Global features extracted from SDF network offer the renderer with shape information, thus helps better reasoning the geometry as well as appearance.
VolSDF \cite{yariv2021volume} further improves the geometry by treating the volume density as a Laplacian cumulative distribution function (CDF) applied to an SDF.
With a well-designed unbiased and occlusion-aware weight function, NeuS \cite{wang2021neus} achieves both quality rendering and fine geometric reconstruction details.

Most of the above approaches rely mainly on pixel values to determine the underlying scene structure with multi-views, thus, view-dependent effects such as reflection on a non-Lambertian surface or inconsistent lighting could easily incur shape-radiance ambiguity \cite{zhang2020nerf++}.
% As a result, NeRF is vulnerable in real scenes with noisy camera poses, sparse views, and varying radiance field.
% % NeRF
% % volume rendering & surface rendering
% Volume-based approaches.
% \cite{mildenhall_2020_nerf,verbin2022refnerf,barron_2021_mip,rasmuson2022addressing,niemeyer2022regnerf}
% aim at novel view synthesis or relighting, do not produce a 3D surface reconstruction of the scene’s geometry.
%\textbf{Radiance decomposition and prior-guided learning.}
% \zbfc{shape-radiance ambiguity.}
It is postulated in \cite{zhang2020nerf++} that NeRF may alleviate the effect of shape-radiance ambiguity by preferring smoothly varying view-dependent radiance field.
%has explored NeRF's success in avoiding shape-radiance ambiguity, with the assumption that radiance for a 3D point varies smoothly if the shape is correctly estimated. 
However, this condition may not be true with glossy surfaces or dynamic illuminations, where the radiance fields are inherently non-smooth. 
Another line of work aims at reconstructing editable representations by combining NeRFs with physics-based rendering techniques \cite{zhang2021nerfactor,boss2021neuralpil,bi2020neural,zhang2021physg,xia2016recovering,yao2022neilf,zhang2022iron,nimier2021material,li2018learning,chen2021dib,bi2020deep,azinovic2019inverse}. 
They decompose scenes into geometric and material factors or even explicitly learn the environmental lighting condition, but are still vulnerable to inputs with complex illumination artifacts.

% \zbfc{and as lighting changes cross viewings, there lacks natural correspondence between viewing directions at image level.}

In order to combat high-frequency factors that may cause difficulties in the reconstruction process, 
several works propose to optimize an extra transient field for each view \cite{martin2021nerfw,zhang2021ners,kuang2022neroic,rasmuson2022addressing,Xian_2021_CVPR,Gafni_2021_CVPR,boss2021nerd,Gao_2021_ICCV,Li_2022_CVPR} so that the shared static structure can be faithfully reconstructed. 
Inspired by NeRF-W \cite{martin2021nerfw} which introduces transient latents to identify and disentangle view-dependent effects in unstructured data, \cite{kuang2022neroic} further employs physics-based rendering techniques \cite{zhang2021nerfactor} for material and normal estimation of an object with internet images. 
Other methods employ guided-optimization to leverage external signals for better geometry  \cite{neff2021donerf,roessle2022densedepthprior,azinovic2022neuralrgbd,wang2022neuralroom,wei2021nerfingmvs,deng2022depthsupervised}. For example, \cite{wei2021nerfingmvs} first learns a monocular depth network on sparse structure-from-motion reconstruction and then utilizes the adapted depth priors to enhance the sampling of the volume rendering procedure. 
Depth oracle \cite{neff2021donerf} also provides useful information for accelerated convergence while facilitating sparse view reconstruction \cite{deng2022depthsupervised}.
% Geo-NeuS also based on the combination of volume and surface rendering in NeuS. \zbf{They leverage explicit supervision on sdf network by sparse points as well as a multi-view photometric consistency constraint to ensure a smooth reconstruction for texture-less region while capture the complex geometric details with strong textures.}
%Both ways are to achieve consistent matching among views, moreover, ensure geometry-consistent reconstruction.

To maximize practical usefulness, we choose not to assume external depth guidance. Instead, our method self-distills invariant features from the radiance field and encodes them back into a neural feature field to normalize the view-dependence.

\section{Method}
\label{sec:method}

We aim at reconstructing the surface (geometry) of a scene with a set of posed images $\{I_k, p_k\}$ captured under dynamic lighting condition within the volume rendering framework \cite{mildenhall_2020_nerf}.
More explicitly, we assume that the light field associated with each image is {\it not} constant.
Without the assumption of a stable lighting, there exists a varying degree of freedom where the radiance of a surface point can change with respect to the viewing direction.
Here we denote this phenomenon as {\it directional view-dependence}.
Please note that we can also observe view-dependence due to non-Lambertian surface, however, by assuming non-constant light field, our method is now put into a more general setup where accurate geometry is desired despite inconsistent radiance. 

In the following, we first analyze how the minimality of the shape-radiance ambiguity varies according to the level of directional view-dependence.  
%we first analyze the difference in optimizing processes while the color of surface points vary frequently as view change, due to a dynamic light field. 
We then introduce a view-dependence normalization method that aligns the minimality of the shape-radiance ambiguity through self-distillation of the encoded invariances in the scene representation.
%Then we introduce our illumination-invariant supervision to help yielding better geometry under both stable and diverse lighting. 
%Finally, we explain our overall training scheme with the modified loss function.
Finally, we show how the image reconstruction and self-distillation for view-dependence normalization can be jointly trained to achieve accurate geometry estimation under non-Lambertian and dynamic lighting conditions.

\begin{figure}[!t]
    \centering
    \includegraphics[width=\linewidth]{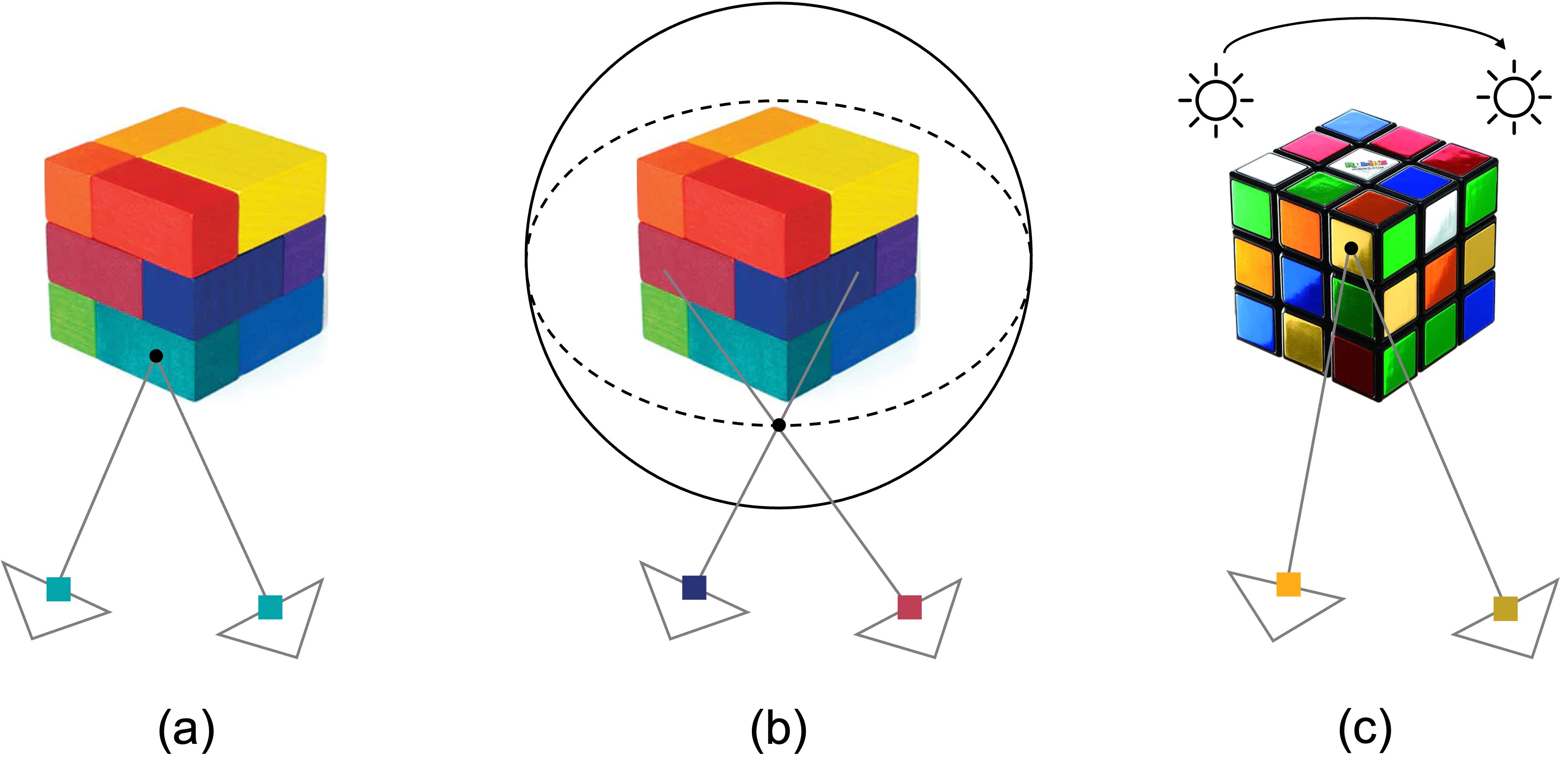}
    \caption{(a): When the geometry is correct, observations (the projections on the 2D image plane represented by squares) of the same surface point from different viewpoints are similar;
    (b): When the geometry is {\it incorrect}, i.e., a cube is reconstructed as a sphere, the radiance of a surface point (dot on the sphere) can exhibit large directional view-dependence. However, as long as the view-dependent radiance function has enough capacity ($c$ in Eq.~\eqref{eq:volume_rend}), volume rendering of the wrong geometry can still achieve small photometric reconstruction error. Thus, one should constrain the view-dependent capacity of the radiance function to avoid overfitting; (c): On the other hand, when the surface is non-Lambertian or the light field is unstable, 
    %{\it even if} the geometry is {\it correct}, the projections of the same point can still be view-dependent.  
    one should {\it not} over-constrain the view-dependent capacity; otherwise, the geometry may be traded for photometric reconstruction quality.}
    \label{fig:shape-radiance-ambi}
\end{figure}

\subsection{Volume rendering with shape-radiance ambiguity and directional view-dependence}

To account for view-dependence, the radiance of a certain point $x\in\mathbb{R}^3$ in the scene is the output of a function: $c: \mathbb{R}^3\times\mathbb{S}^2\rightarrow\mathbb{R}^3$, which maps the point coordinate $x$ and a viewing direction $\mathbf{v}\in\mathbb{S}^2$ to an RGB value.
Given a certain pixel $\mathbf{p}$ on the image plane defined by a camera center $\mathbf{o}$, 
we can leverage the commonly used volume rendering formulation to compute its pixel value as:
\begin{equation}
    C(\mathbf{o},\mathbf{v})
    =\int\limits_0^{+\infty} w(t)c(\mathbf{p}(t),\mathbf{v})\mathrm{d}t
    \label{eq:volume_rend}
\end{equation}
where $\{\mathbf{p}(t)=\mathbf{o}+t\mathbf{v}\mid t\geq 0\}$ is the (outward) ray passing through the camera center $\mathbf{o}$ and the pixel $\mathbf{p}$.
And $w$ is a weight function that depends on the scene geometry, and ideally, we like it to peak at the surface point.

\begin{figure}[!ht]
    \begin{subfigure}{1.0\linewidth}
        \centering
        \includegraphics[width=\linewidth]{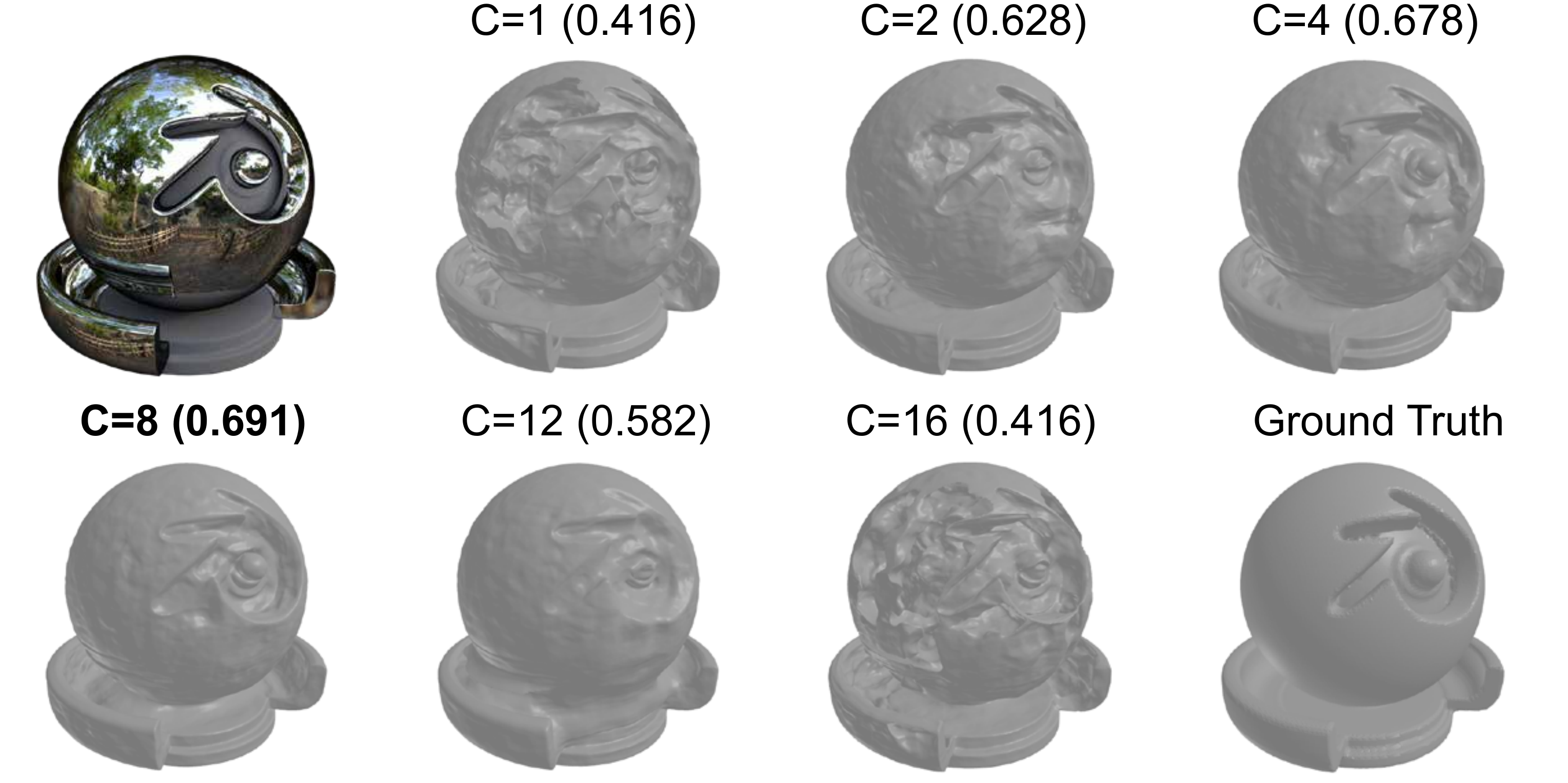}
        \caption{Geometric quality (with a fixed set of posed images) varies according to the capacity of the view-dependent color function. 
        The scalar in the bracket is the intersection-over-union score, and C is the number of layers in the view-dependent color MLP.
        Bold indicates the optimal capacity. Here the object has a high directional view-dependence.}
        \vspace{0.4cm}
        \label{fig:ball}
    \end{subfigure}
    % \begin{subfigure}{1.0\linewidth}
    %     \centering
    %     \includegraphics[width=0.94\linewidth]{figures/shape-capacity.png}
    %     \caption{\zbf{Shape-randiance ambiguity}}
    %     \label{shape-capacity}
    % \end{subfigure}
    \begin{subfigure}{1.0\linewidth}
        \centering
        \includegraphics[width=\linewidth]{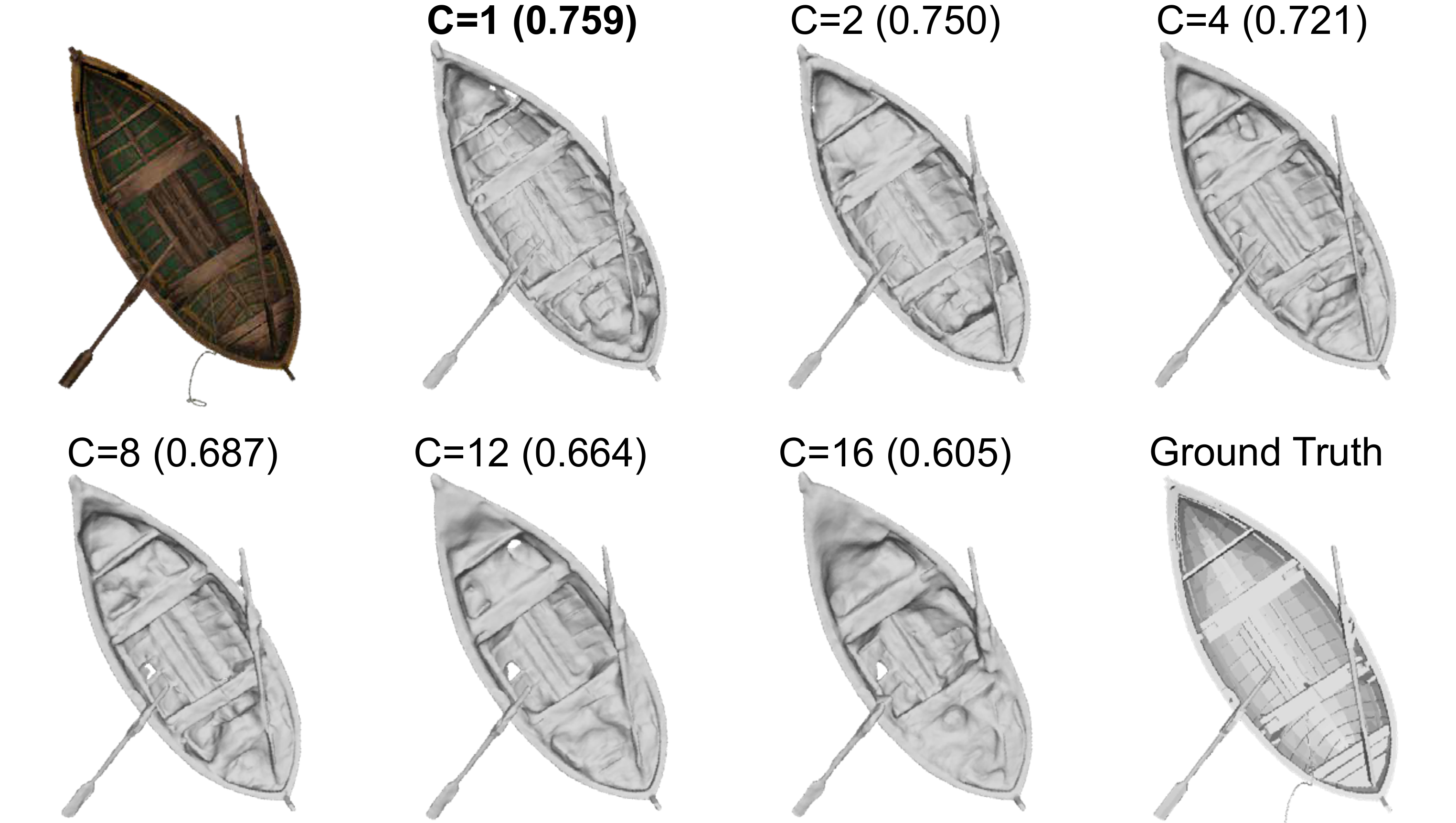}
        \caption{Similar to the above example, but now the object has a low directional view-dependence, and the optimal result is achieved at a small capacity.}
        \vspace{0.4cm}
        \label{fig:boat}
    \end{subfigure}
    \begin{subfigure}{0.49\linewidth}
        \includegraphics[width=\linewidth]{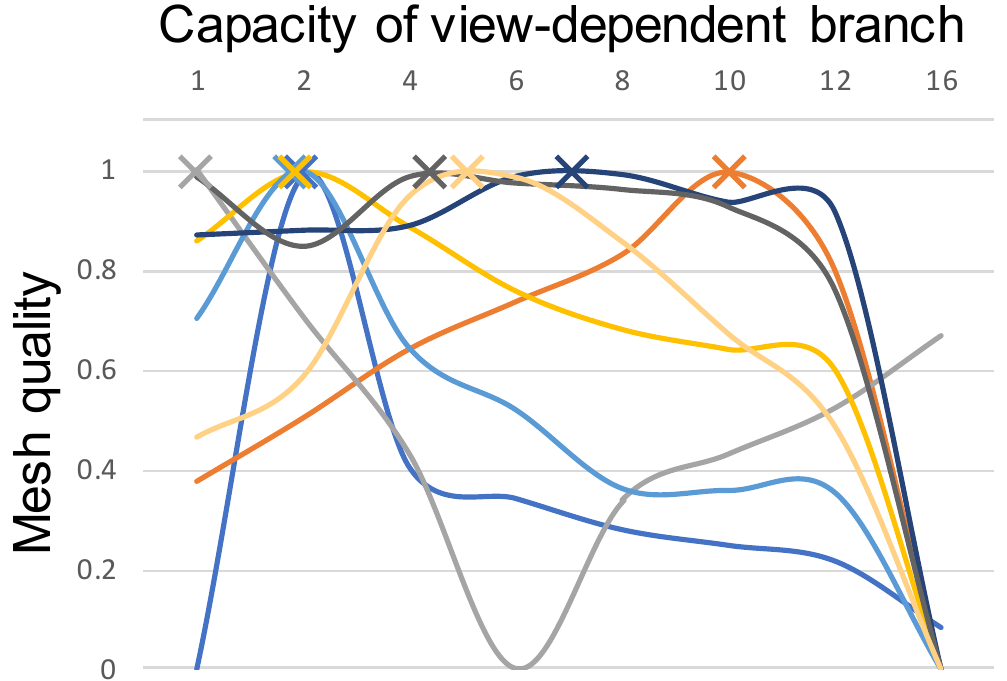}
        \caption{NeuS}
        \label{fig:neus-f}
    \end{subfigure}
    \begin{subfigure}{0.49\linewidth}
        \includegraphics[width=\linewidth]{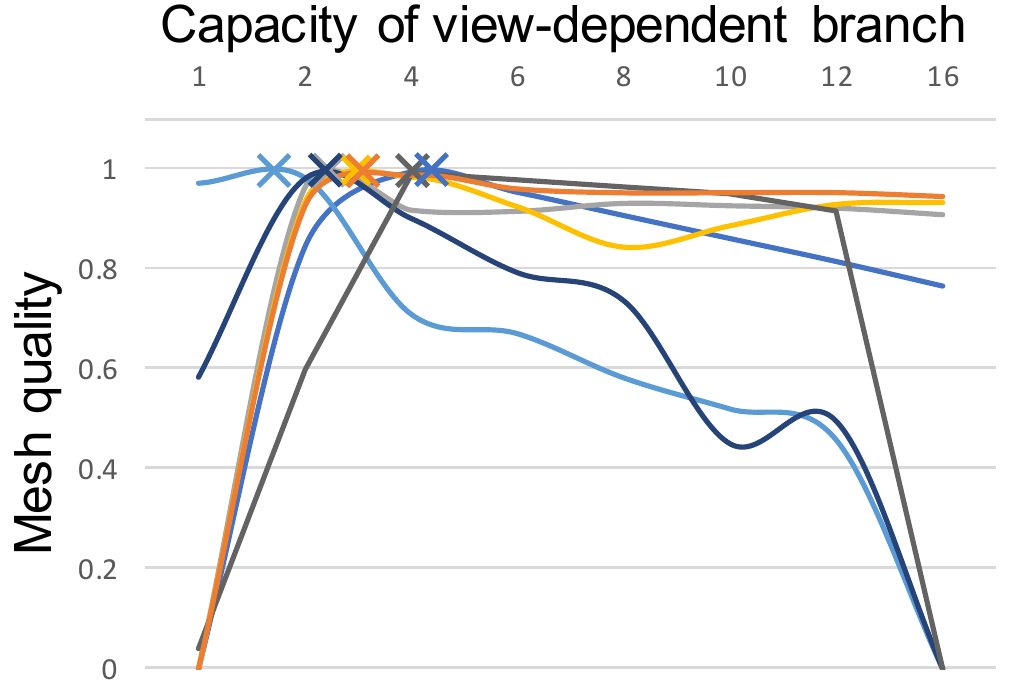}
        \caption{Ours}
        \label{fig:ours-f}
    \end{subfigure}
    \caption{
    In {\bf (a) and (b)}, we show two examples with different levels of directional view-dependence.
    As observed, the geometric quality of both depends on the capacity of the view-dependent color function,
    where the one with higher view-dependence achieves its optimum at a relatively larger capacity.
    Moreover, both decrease when the capacity gets larger than enough, entering the shape-radiance ambiguity zone.
    {\bf (c):} For each object, we traverse the view-dependent color function's capacity in \cite{wang2021neus} and plot the curve of the geometric quality. We can see that the coupling effect illustrated above is statistically valid,
    and there is no consensus on a capacity to achieve the optimum performance for all scenes. 
    {\bf (d):} When varying the view-dependent branches in the proposed framework (Fig.~\ref{fig:pipeline}), we observe that the optimal capacity for all is consistent, verifying the effectiveness of the proposed directional view-dependence normalization. Note that the curves are scaled against the maximum for better visualization.}
    \vspace{-0.2cm}
    \label{fig:coupling}
\end{figure}

Now we introduce our choice of the weight function beforehand to set a concrete ground for the consecutive analysis.
However, note that our analysis of the coupling between the shape-radiance ambiguity and directional view-dependence of radiance {\it does not} rely on the exact choice of the weight function as long as it encourages maximum importance in the vicinity of surface points.
Specifically, we adopt the weighting scheme of \cite{wang2021neus} as in the following:
\begin{equation}
    w(t) = T(t)\rho(t)
    \label{eq:weight}
\end{equation}
where $\rho(t)$ is the density of a point on a ray parameterized by $t$, and $T(t)=\mathrm{exp}(-\int_0^t\rho(\tau)\mathrm{d}\tau)$ is the accumulated transmittance.
Particularly,
we have 
%the $T(t)$ denotes the accumulated transmittance along the ray, and $\rho(t)$ is the underlying opaque density:
%\begin{equation}
%    T(t)=exp(-\int\limits{0}^{t}\rho(u)\;\mathrm{d}u)
%    =\Phi_{s}(sdf(\mathbf{p}(t))),
%    \label{eq:T(t)}
%\end{equation}
\begin{equation}
    \rho(t)=\mathrm{max}(\frac{-\frac{\mathrm{d}\Phi_s}{\mathrm{d}t}(f(\mathbf{p}(t)))}{\Phi_s(f(\mathbf{p}(t)))}, 0)
    \label{eq:defrho}
\end{equation}
where $f: \mathbb{R}^3\rightarrow\mathbb{R}$ is a signed distance function parameterized by a neural network, and $\Phi_s$ is the Sigmoid function.
Next, we detail the relationship between shape-radiance ambiguity and directional view-dependence in the context of volume rendering descried by Eq.~\eqref{eq:volume_rend}.

{\bf Shape-radiance ambiguity} in our case refers to the degeneration of geometry when the posed images can be perfectly reconstructed through volume rendering.  In Fig.~\ref{fig:shape-radiance-ambi} (a), we show a Lambertian rubik, where different projections of the same surface point are constant if the geometry is correctly estimated.
However, as soon as the capacity of the view-dependent color function (multi-layer perceptron $c(\mathbf{p}(t),\mathbf{v})$ in Eq.~\eqref{eq:volume_rend}) becomes large enough,
the posed images of the rubik can still be reconstructed even if the geometry is completely wrong.
As shown in Fig.~\ref{fig:shape-radiance-ambi} (b), 
one can think of the sphere around the rubik as an LED screen, which can synthesize the image of an arbitrary camera pose by adjusting the color of each unit.
In this case, the view-dependent color function's capacity is approaching infinity.
Thus, to ensure the quality of the learned geometry, we need to impose a constraint on the color function.

However, {\bf directional view-dependence} is universal as there exist many non-Lambertian surfaces in the real world; moreover, there is no guarantee that the light field is static, 
especially when there is a need for active light sources, for example, in dim environments.
And usually, we have a mixture of them, as shown in Fig.~\ref{fig:shape-radiance-ambi} (c).
In order to have a good geometry through volume rendering,
we may want to increase the capacity of the view-dependent color function so that
the photometric reconstruction objective {\it does not} distort the geometry (through $w(t)$ in Eq.~\eqref{eq:volume_rend}).

\begin{figure*}
    \centering
    \includegraphics[width=\linewidth]{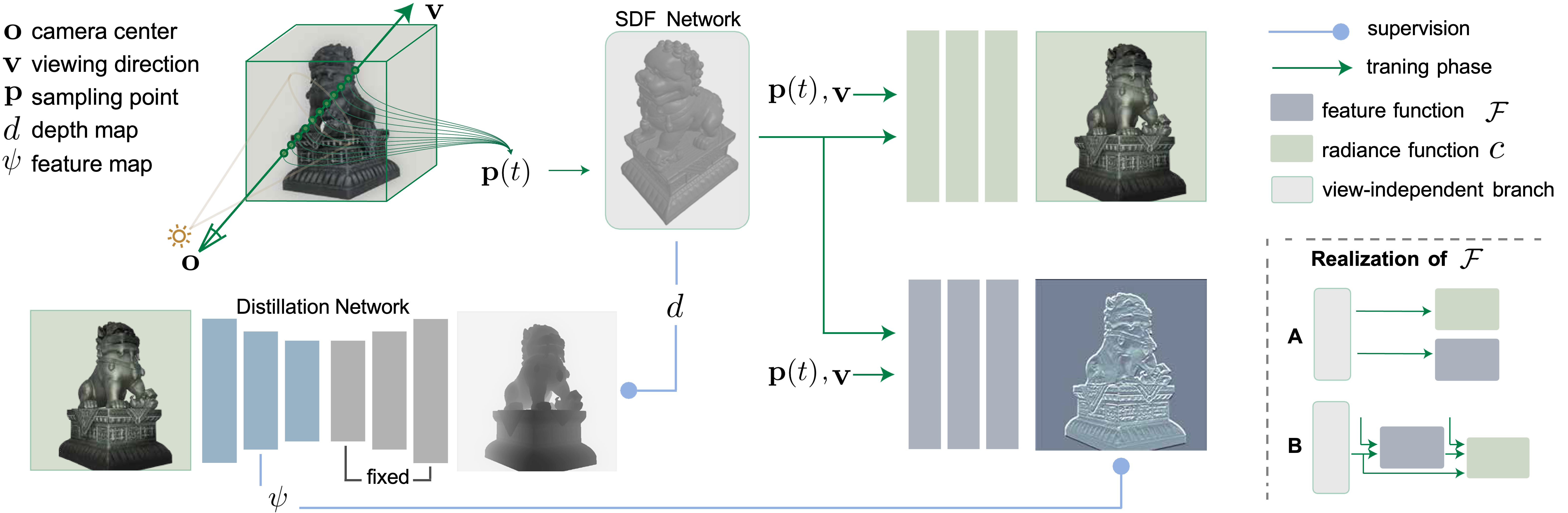}
    \caption{Overview of our method. When reconstructing the scene from a set of posed images captured under dynamic lighting conditions, the proposed view-dependence normalization aligns the minimality of the shape-radiance ambiguity for different scenes by self-distilling depth features into a neural feature field. Note that the depth supervision for the distillation network comes directly from the SDF network in the architecture instead of external depth oracles. Two ways for incorporating the feature branch into the neural radiance field are shown in the bottom right, where A represents the parallel scheme in the left pipeline, and B indicates a scheme that makes the color function dependent on the feature branch.}
    \label{fig:pipeline}
\end{figure*}

Thus, we can see a {\bf tight coupling} between shape-radiance ambiguity and directional view dependence in the context of volume rendering.
To account for view dependence, the capacity of the (directional) color function needs to be increased; 
yet, we {\it do not} want the capacity to pass what is enough for rendering the exhibited variations in the radiance.
Otherwise, the shape-radiance ambiguity kicks in and starts to decrease the estimated geometry's accuracy.
We can check this phenomenon in Fig.~\ref{fig:coupling} (a) and (b), where the metallic ball has a higher view-dependence radiance variation than the boat.
We train NeuS \cite{wang2021neus} for each of them, but with various numbers of MLPs for the view-dependent color branch.
As the capacity increases, the geometric quality of the reconstructed ball also increases until it hits a threshold (layer number of MLPs equal to eight in Fig.~\ref{fig:coupling} (a)).
Similarly, the optimal performance for the boat is achieved with one layer of MLP, and then starts to decrease (Fig.~\ref{fig:coupling} (b)).

These examples clearly convey two messages: 1) The shape-radiance ambiguity always exists as long as the view-dependent color function has a large enough capacity;
2) We can alleviate the shape-radiance ambiguity by reducing the capacity until an optimal performance, but no more.
Given the second, we could tune the view-dependent color function's capacity to achieve the best geometric quality for each scene.
{\it However,} this requires a time-consuming training process and relies on a reference geometry to tell the optimum.
As evidenced in Fig.~\ref{fig:coupling} (c),
the optimal capacity for each scene (crosses on the performance-capacity curves) varies depending on the level of view-dependence of the scene, i.e., higher view-dependence entails a larger optimal capacity.

Instead of tuning the capacity for each scene,
we ask whether it is possible to align the degree of view-dependent radiance variation for all scenes.
By doing so, a single optimal capacity should work across scenes and achieves the best overall geometry.
Next, we describe how this can be achieved without controlling the surface material or the lighting condition, as shown in Fig.~\ref{fig:coupling} (d).

%\cref{fig:pipeline} shows an overview of our approach. Given a set of posed images $\{I_k\}$ of a 3D object, our goal is to reconstruct the surface $S$ of it. Following \cite{yariv2020idr,wang2021neus}, we adopt an approach that performs surface rendering by training in a volume rendering scheme. Geometry of the scene is represented by a signed distance function (SDF), where foreground object could be extracted by searching the zero-level set. Previous methods \zbf{[][]} have provided solutions to learn high-quality geometry from multi-view images rendered or taken under scenes with static illuminations, but tend to fail when it comes to dynamic illuminations. 

%In applying of these approaches, shape-radiance ambiguity studied in \cite{zhang2020nerf++} is an inescapable problem due to the lack of explicit 3D constraints, especially when sparsity of input images raises. While network optimized with rendering loss only, the constraints could be satisfied by modifying either the geometry branch $\mathcal{G}$ or the view-dependent radiance field, so that there is no guarantee for a correct geometry. In another aspect, when our goal turns to precise geometry, the assumption in \cite{zhang2020nerf++} is of little use and intricately related factors would make the situation much fancier.\yanchao{what is the point that you want to say here?}

\subsection{Directional view-dependence normalization}
\label{sec:vdn}

%In neural rendering, we could decompose the networks into two parts, the first to capture the geometry information (geometry branch $\mathcal{G}$) and the second to extract a correspond radiance field (beta branch $\mathcal{B}$). Geometry is represented by density in volume rendering, while by SDF in surface rendering. The radiance field, generated by concatenating geometry information with view directions is always represented by the pixels' colors in images. As analyzed in \cite{zhang2020nerf++}, the correct shape would cause a smoother surface radiance field and in contrast incorrect geometry would force the radiance field to have higher intrinsic complexity (i.e., much higher frequencies). NeRF's MLP structure, so as other rendering methods that assign low capacity to the view-dependent color branch, encocdes the favor of smooth surface reflectance function and regularize view-dependent colors, so the shape-radiance could be avoided in static scene. 

% \zbf{\cref{fig:rolleiflex_psnr} also shows us that with the increasing of the beta branch's capacity, the shape-radiance ambiguity exacerbates: qualities of rendered images of both training views and test views decline to variant degrees.}
% In the following, we analyze the phenomenon by exploring the learned geometry's quality in extreme situations. Quantitative results for NeuS are showed in \cref{fig:neus-f} for geometry and \cref{fig:rolleiflex_psnr} for rendering, we will discuss about it later.

% \zbfc{(nerf++ focus on rendering quality, we should bring it to analysis on geometry.)}

Ideally, we like the scene to be perfectly Lambertian and the light field stable during the image capturing.
In other words, we want the radiance of a specific point on the surface to be viewpoint invariant (Lambertian) and illumination invariant (stable lighting).
Note that since an active light source usually changes according to the vantage point of the observer, we can think of the two as viewpoint-induced variations.
We introduce our directional view-dependence normalization method that helps achieve the desired invariances and the joint training scheme for quality geometry via volume rendering (Fig.~\ref{fig:pipeline}).

In contrast to \cite{wei2021nerfingmvs,rasmuson2022addressing} that alleviate the effect of shape-radiance ambiguity by either external geometry priors or separate modeling of view-dependent and Lambertian components,
the proposed view-dependence normalization works directly with the original neural radiance field through view-invariance distillation.
More explicitly, we instantiate a distillation network $\psi$, which takes in a rendered image $I^r_k$ and predicts the corresponding rendered depth:
\begin{equation}
    d^r_k = \hat{d^r_k} = \psi(I^r_k)
    \label{eq:rendered_depth}
\end{equation}
where $d^r_k$ is the projection of the zero-level set $S=\{x\in\mathbb{R}^3\mid f(x)=0\}$ under the same pose of $I^r_k$.
Also denote $\psi^l(I^r_k)$ as the feature map from the $l$-th layer of $\psi$, with $\hat{d^r_k}=\psi^L(I^r_k)$.
Explicitly, we adopt \cite{ramamonjisoa2021wavelet}, which employs \cite{huang2017densely} as its encoder, to serve as the distillation network $\psi$.
Then, we instantiate a feature function $\mathcal{F}$, in parallel to the radiance function $c$ (Fig.~\ref{fig:pipeline}), and denote the corresponding volume rendering as:
\begin{equation}
    \Psi(\mathbf{o},\mathbf{v})
    =\int\limits_0^{+\infty} w(t)\mathcal{F}(\mathbf{p}(t),\mathbf{v})\mathrm{d}t
    \label{eq:volume_rend_feat}
\end{equation}
where ($\mathbf{o},\mathbf{v}$) specify a ray that is parameterized by $t$. 

The {\it key} of view-dependence normalization is to dilute the variation in colors with the prediction of a more view-invariant signal, namely, we set:
\begin{equation}
    \Psi(\mathbf{o},\mathbf{v}) = \bar{\psi}^l(I^r_k)[\mathbf{p}]
\end{equation}
where $\bar{\psi}^l(I^r_k)$ is the normalized feature from the $l$-th layer of the depth prediction network $\psi$.
Since depth prediction requires illumination invariance (luminance does not change the geometry), features useful for this task should be stable to light field changes.
Similarly, depth features shall encode a certain level of viewpoint invariance, e.g., 
in-plane translation and rotation of the camera do not alter the depth of the same point.
Moreover, normalizing the depth feature help throw away depth variations caused by out-of-plane perturbations of camera pose,
which further enhances the viewpoint invariance of the features.

On the other hand, features encoded by $\mathcal{F}$ should maintain discriminability modulo view-dependence so that geometric details are reconstructable.
One can achieve this by choosing a relatively smaller $l$ for $\bar{\psi}^l(I^r_k)$, which we will ablation study as an operational hyperparameter.
By learning the feature field in addition to radiance, 
viewpoint variations in color shall be compensated by the view-invariance in the distilled features, thus normalizing the view dependence.
Now we detail the joint training objectives.

\subsection{Network training}

The proposed joint training with view-dependence normalization minimizes both the photometric and self-distilled feature reconstruction errors.
%We minimize the difference between both the rendered colors and the ground truth colors, and the rendered features and the feature from depth net. 
Denote the input image as $I_k$, the batch size as $n$ and the number of point samples per ray as $m$,
%corresponding to k and i as their indexes respectively. 
the joint training objective is:
% Besides, we also utilize the masks for supervision if provided.
\begin{equation}
    \mathcal{L}=\lambda_{color}\mathcal{L}_{color}+\lambda_{vdn}\mathcal{L}_{vdn}+\lambda_{reg}\mathcal{L}_{reg}
    % +\lambda_{mask}\mathcal{L}_{mask}
\end{equation}
where the color loss $\lambda_{color}$ and view-dependence normalization loss $\mathcal{L}_{vdn}$ are defined as:
\begin{equation}
    \mathcal{L}_{color}=\frac{1}{n}\sum\limits_{i}|I^r_{k,i}-I_{k,i}|
\end{equation}
\begin{equation}
    \mathcal{L}_{vdn}=\frac{1}{n}\sum\limits_{i}|\Psi(\mathbf{o}_i,\mathbf{v}_i) - \bar{\psi}^l(I_k)[\mathbf{p}_i]|
\end{equation}
%where $l$ is set to be 1.
Here, we use L1 loss as in \cite{yariv2020idr,wang2021neus}.
%The color loss shall ensures that network find a reasonable point leading to convergence at early stage. 
And $\mathcal{L}_{reg}$ is an Eikonal term \cite{gropp2020implicit} regularizing the gradients $\nabla$ of the SDF network:
\begin{equation}
    \mathcal{L}_{reg}=\frac{1}{mn}\sum\limits_{i,j}(||\nabla f(\mathbf{p}_{i,j}))||-1)^2
\end{equation}
Also as in \cite{wang2021neus}, when masks are available, the color loss and mask loss are defined as:
% integrate masks (different from NeuS, we separately restrain foreground pixels which are to calculate $\mathcal{L}_{color}$ and  the distance between the weights' distribution and ground truth masks weight. This is in case for those vague foreground pixels.)
\begin{equation}
    \mathcal{L}_{color}=\frac{1}{n}\sum\limits_{i}|I^r_{k,i}-I_{k,i}|\cdot M_{k,i}
\end{equation}
\begin{equation}
    \mathcal{L}_{msk}=BCE(M_k, \hat{M_k})
\end{equation}
where $M_k$ is the foreground mask for image $I_k$, and $\hat{M_k}=\sum\nolimits_{j=1}^m w_{k,j}$ is the sum of weights along the camera ray.
{\bf Camera pose optimization}. Similar to \cite{wang_2021_nerfmm}, we also treat camera poses and intrinsics as learnable parameters, and jointly optimize them when it is needed. 
For more details
%on
about
this part please refer to the appendix.
% For example, COLMAP, a widely used tool to estimate coarse camera pose, always give a camera pose which is not that precise, and we can optimize the pose in order to get better geometry.

\section{Experiments}

\begin{table*}[]
    \centering
    \resizebox{\linewidth}{!}
    {
    \begin{tabular}{@{}l||cc||cccc||cccc||ccc@{}}
      \toprule
        % Method & \footnotesize COLMAP & \footnotesize Plenoxels & \footnotesize NeRF & \footnotesize NeRF-W & \footnotesize NeROIC & \footnotesize RefNeRF & \footnotesize VolSDF & \footnotesize NeuS & \footnotesize NeuS-S & \footnotesize GeoNeuS & \footnotesize VolSDF+$\mathcal{F}$ & \footnotesize NeuS+$\mathcal{F}$ & \footnotesize NeuS-S$+\mathcal{F}$ \\
        Method & \small COLMAP & \small Plenoxels & \small NeRF  & \small NeRF-W & \small NeROIC & \small RefNeRF & \small VolSDF & \small NeuS & \small Geo-A & \small GeoNeuS & \small VolSDF+$\mathcal{F}$ & \small Geo-A$+\mathcal{F}$ & \small Ours
        \\
        %  &  \\
        %  & 
      \midrule
        IoU $\uparrow$ & 0.607 & 0.479 & 0.481 & 0.586 & 0.501 & 0.442 & 0.635 & 0.639 & 0.600 & 0.517 & 0.675 & \underline{0.692} & \textbf{0.708} \\
        L1 CD $\downarrow$ & 0.587 & 0.760 & 0.938 & 0.672 & 0.999 & 1.750 & 0.530 & 0.563 & 0.638 & 0.781 & 0.465 & \underline{0.434} & \textbf{0.397} \\
        L2 CD $\downarrow$ & 2.568 & 1.897 & 3.263 & 1.549 & 3.938 & 15.152 & 1.223 & 1.461 & 2.416 & 2.873 & 1.063 & \underline{0.898} & \textbf{0.827} \\
        NC $\uparrow$ & 0.785 & 0.636 & 0.654 & 0.725 & 0.704 & 0.687 & 0.829 & 0.831 & 0.808 & 0.768 & \underline{0.837} & 0.829 & \textbf{0.845} \\
        f-score $\uparrow$ & 0.790 & 0.577 & 0.552 & 0.668 & 0.568 & 0.521 & 0.760 & 0.762 & 0.757 & 0.627 & 0.798 & \underline{0.834} & \textbf{0.854} \\
      \bottomrule
    \end{tabular}
    }
    \vspace{-0.2cm}
    \caption{Quantitative comparison on our dataset with non-learning-based approaches: COLMAP \cite{schonberger2016pixelwise}, Plenoxels \cite{fridovich2022plenoxels}; volume-based approaches: NeRF \cite{mildenhall_2020_nerf}, NeRF-W \cite{martin2021nerfw}, NeROIC \cite{kuang2022neroic}, RefNeRF \cite{verbin2022refnerf}; and SDF-based methods: VolSDF \cite{yariv2021volume}, NeuS \cite{wang2021neus}, Geo-A \& GeoNeuS \cite{fu2022geo}.
    Note that Geo-A refers to a simplified version of GeoNeuS \cite{fu2022geo} (named Model-A in their ablations), where the SDF is supervised by sparse points from COLMAP. 
    We apply the proposed view-dependence normalization to VolSDF (VolSDF+$\mathcal{F}$), Geo-A (Geo-A$+\mathcal{F}$), and NeuS (Ours).
    The effectiveness is observed in all the variants of our method.
    Moreover, ours achieves the best on all five metrics of the learned geometry: Intersection-over-Union (IoU), L1/L2 Chamfer Distance (CD), Normal Consistency (NC), and f-score.}
    \vspace{-0.4cm}
    \label{tab:all-baselines}
\end{table*}

\begin{figure}
    \centering
    \includegraphics[width=\linewidth]{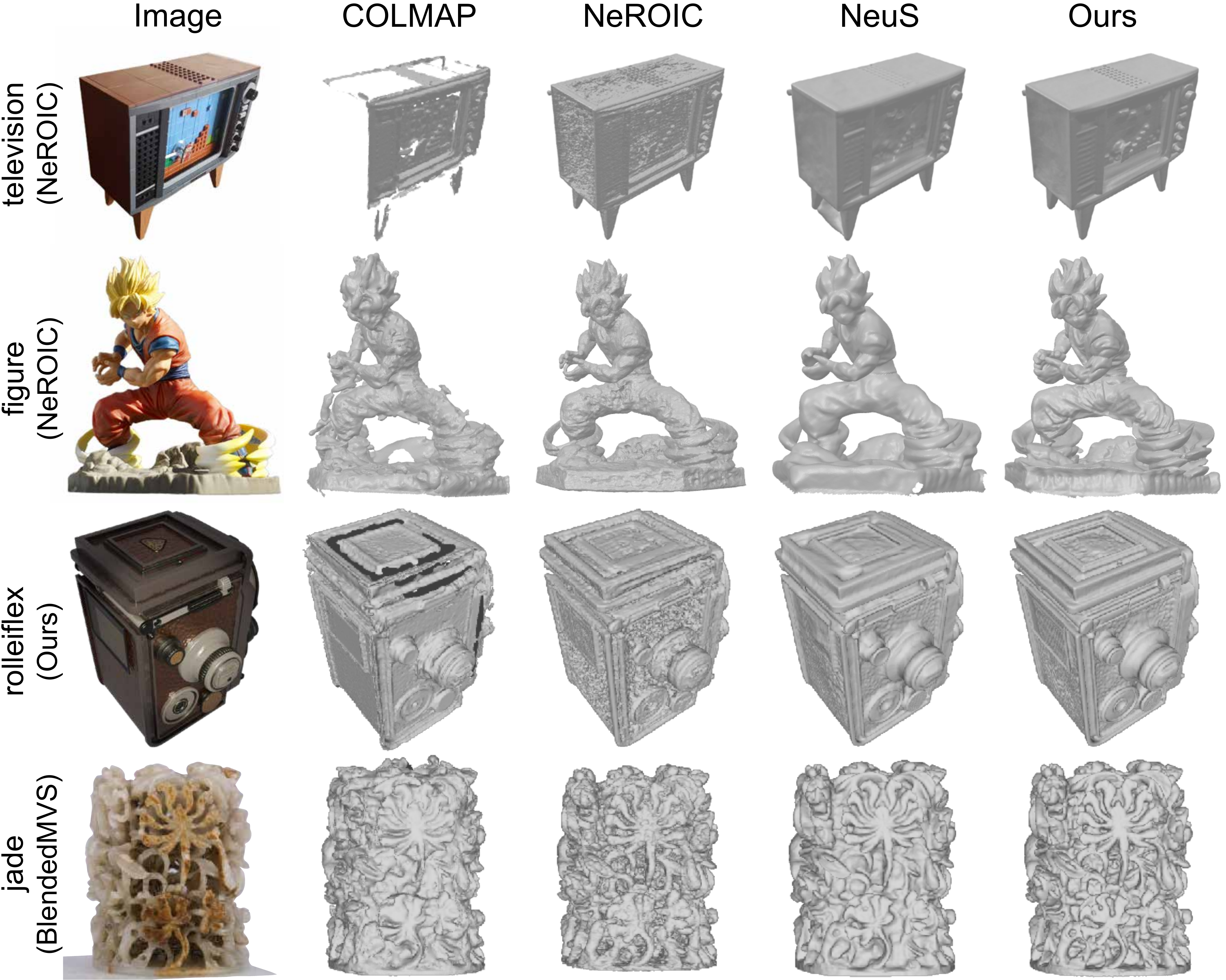}
    \caption{Qualitative comparison with COLMAP \cite{schonberger2016pixelwise}, NeROIC \cite{kuang2022neroic}, and NeuS \cite{wang2021neus}, which are representatives of different categories described in Tab.~\ref{tab:all-baselines}.
    %Qualitative results on NeROIC \cite{kuang2022neroic}, BlendedMVS \cite{yao2020blendedmvs} and our dataset. We respectively select COLMAP \cite{schonberger2016pixelwise}, NeROIC \cite{kuang2022neroic}, and NeuS \cite{wang2021neus} as the representative method for each of non-learning-based methods, volume-based methods, and SDF-based methods. 
    Our model retains accurate geometry with more details and less artifacts.}
    \vspace{-0.4cm}
    \label{fig:allmethods}
\end{figure}

% setup for training (or Implementation details)
% \textbf{Implementation details.} We implement our method on top of NeuS \cite{wang2021neus}, with our feature function $\mathcal{F}$ in parallel to the radiance function $c$. They share the same architecture of 4-layer MLP with hidden dimension of 256.

\textbf{Implementation details.} We implement our method mainly based on NeuS \cite{wang2021neus}. The feature function $\mathcal{F}$ and the radiance function $c$ have similar architectures, i.e., 4-layer MLP with a hidden dimension of 256.
% As a default setting, we follow the network architecture in NeuS \cite{wang2021neus} and add our feature branch after sdf network. The feature branch shares the same capacity as an original color branch, i.e., a 4-layer ReLU MLP with a sigmoid activate function before output.
We follow the hierarchical sampling strategy in NeuS \cite{wang2021neus} and set the batch size to 512.
We also adopt \textit{WaveletMonodepth} \cite{ramamonjisoa2021wavelet} with a DenseNet161 backbone \cite{huang2019convolutional,huang2017densely} as our distillation network, and pretrain it on in-domain data (different from training and test scenes) for a few epochs to ensure efficient convergence during the joint training stage. 
As an operational choice, we conduct our experiments with depth features from the distillation encoder's first Conv block ($l$=1). 
Note that we can optionally choose not to update the distillation network if the discrepancy between its prediction and the self-supervision is small.
This can help reduce the training time when we have to train many scenes.
The training is performed on a single NVIDIA RTX3090 GPU and takes 300K steps to converge in about 11 hours.

% datasets for evaluation, baselines
% evaluation metrics

\textbf{Baselines.}
We evaluate the proposed method against non-learning-based methods \cite{schonberger2016pixelwise,fridovich2022plenoxels}, volume-based methods \cite{mildenhall_2020_nerf,martin2021nerfw,kuang2022neroic,verbin2022refnerf} , and SDF-based methods \cite{yariv2021volume,wang2021neus,fu2022geo}. 
Meshes for COLMAP \cite{schonberger2016pixelwise} are extracted using screened Poisson Surface reconstruction \cite{kazhdan2006poisson} with trimming value 7.
As a multi-stage method, NeROIC \cite{kuang2022neroic} %first
infers the surface and normals in the first stage, then estimates material properties and ambient illumination with fixed geometry in the second stage. 
We only train it with the first stage to extract the geometry for evaluation.
GeoNeuS \cite{fu2022geo} leverages sparse 3D points from Structure-From-Motion and photometric consistency in Multi-View-Stereo to optimize via SDF loss and photometric consistency loss.
We follow the descriptions in their paper and reimplement their method for evaluation.
% and extract the meshes by marching cube with threshold 50.
% SDF-based methods \cite{wang2021neus}\cite{yariv2021volume} are trained without mask and the meshes are extracted at resolution of 512. 

\textbf{Datasets.}
% As our target is scenes reconstruction under unstructured lighting, our evaluations uses . Our datasets are from three different sources: 
% We perform experiments on synthetic renderings of objects generated by ourselves.
Due to the lack of ground-truth geometry and controls of illumination in existing sequences, we create a new benchmark with synthetic renderings of eight challenging scenes.
Especially, we let the light source changes its location accordingly to the camera poses.
By this, we can simulate different levels of view variations beyond reflections.
In each scene, we randomly select 30 views as training data.
We also show the results on the released data with unconstrained illumination from NeROIC \cite{kuang2022neroic}, as well as the data from DTU \cite{jensen2014large} and  BlendedMVS \cite{yao2020blendedmvs} with static illumination. To test the capability in alleviating shape-radiance ambiguity by the proposed view-dependence normalization in practical applications, we also conduct experiments on two real-world captures, i.e., one is an intra-oral scanning, and the other is the underwater dataset AQUALOC \cite{ferrera2019aqualoc}, where dynamic lighting is needed to better perceive the scene.
% self-captured data shown in NeROIC (Milk, Figure, TV), where each object is placed in about 4-6 different scenes. 
% For general comparison, we also test our method on several universal datasets of multi-view reconstruction/rendering, i.e. NeRF dataset, BlendedMVS dataset, following the experiment setting in NeuS.
% dynamic illumination: private dataset, data provided by NeROIC. static illumination: blendedmvs\&dtu dataset, nerf dataset. real data: include intra-oral scanning data, and frames from the AQUALOC Dataset (harbor video sequences)
 
\textbf{Evaluation.}
% We evaluate on the task of multi-view reconstruction: given a trained neural radiance field, we generate a $256^3$ dense grid inside bounding box to query for the density/ SDF value, then extract a mesh by Marching Cubes \cite{lorensen1987marching} and compare it to the ground truth. We report the quantitative results by measuring Intersection-Of-Union (IoU), L1/L2 Chamfer Distance (CD), Normal Consistency (NC), and f-score. Our evaluation code is based on \cite{siddiqui2021retrievalfuse}.
We generate dense grids of the density/SDF value from the trained neural radiance field and extract meshes by Marching Cubes \cite{lorensen1987marching}. Then, We compare the multi-view reconstruction results to the ground truth and report the quantitative results by measuring Intersection-of-Union (IoU), L1/L2 Chamfer Distance (CD), Normal Consistency (NC), and f-score, following \cite{siddiqui2021retrievalfuse}.

\subsection{Comparison}

\begin{table} % [htbp]
  \centering
    \resizebox{\linewidth}{!}
    {
    \begin{tabular}{@{}lcccccc@{}}
    \toprule
      \multirow{2}{*}{Lighting} & \multicolumn{3}{c}{NeuS} & \multicolumn{3}{c}{ Ours} \\
      \cmidrule(lr){2-4} \cmidrule(lr){5-7}
      & IoU $\uparrow$ & fscore $\uparrow$ & L1 CD $\downarrow$ & IoU $\uparrow$ & fscore $\uparrow$ & L1 CD $\downarrow$ \\
    %   & IoU & fscore & L1 CD & IoU & fscore & L1 CD \\
      \midrule
      static illum. & 0.631 & 0.765 & 0.586 & 0.668 & 0.813 & 0.494 \\
      dynamic illum. & 0.574 & 0.720 & 0.644 & 0.658 & 0.808 & 0.496 \\
      \bottomrule
    \end{tabular}
    }
    \vspace{-0.2cm}
    \caption{Comparison with NeuS \cite{wang2021neus} under different lighting conditions. Our method achieves better scores on all three metrics, especially with dynamic lighting. 
    % Our mehthod is less affected by the dynamic illumination, especially in IoU, our method drops by 0.01, while Neus drops by 0.057.
    %   trained with $\mathcal{F}$ performs better in both illuminations while reconstructions of NeuS visibly deteriorate.
    %   Results on 10-object dataset varying in lighting. 'static', 'spot\_0', 'spot\_1' represent scenes rendered under static ambient light, low brightness and higher brightness spotlights respectively.
    }
    \vspace{-0.4cm}
    \label{tab:vary-in-lights}
\end{table}

\begin{figure}
    \centering
    \includegraphics[width=0.96\linewidth]{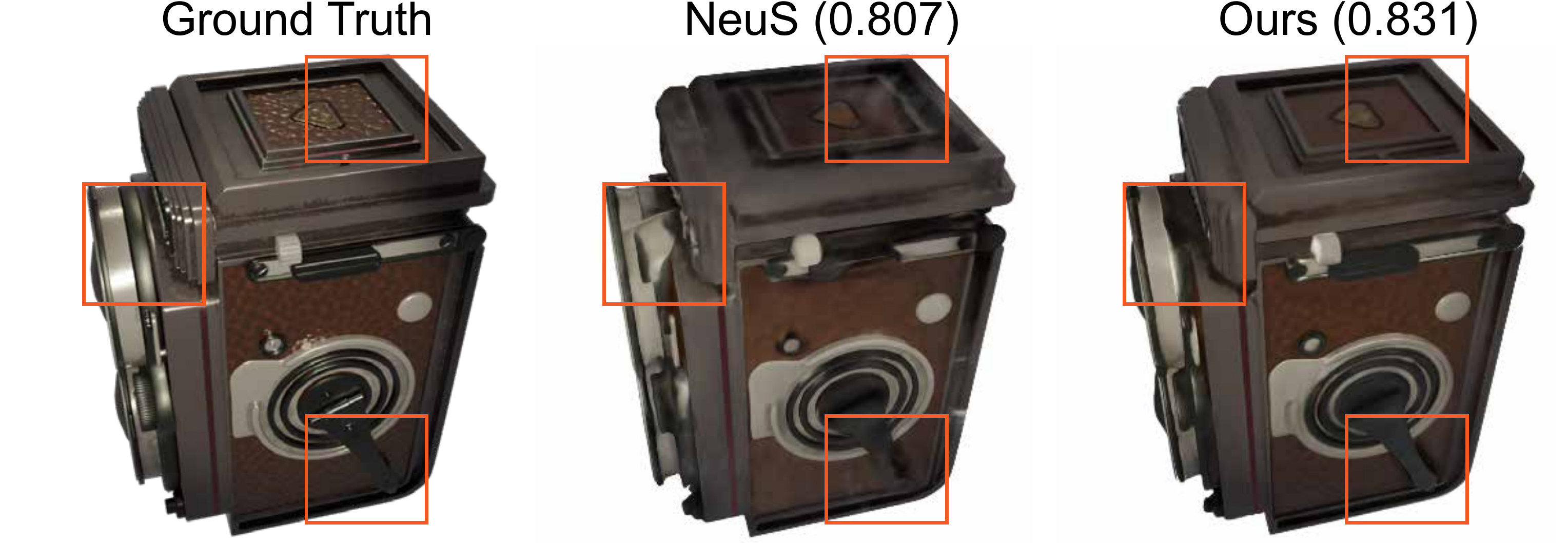}
    \vspace{-0.1cm}
    \caption{Novel view synthesis from NeuS \cite{wang2021neus} and our model. Scalars in brackets are structural similarities (SSIM). Our method learns better geometry and enhances the radiance field, as demonstrated in the improved rendering details.
    }
    \vspace{-0.5cm}
    \label{fig:rolleiflex_rendered}
\end{figure}

% To measure the capability of handling various scenes, we test on 8 challenging scenes captured with a spotlight tracking to camera.
\cref{tab:all-baselines} shows quantitative evaluations for all models on our dynamic-lighting scenes.
SDF-based methods \cite{wang2021neus,yariv2021volume} outperform volume-based methods \cite{mildenhall_2020_nerf, kuang2022neroic, verbin2022refnerf} by an average margin of 0.1 in IoU and 0.17 in f-score. As observed in \cref{fig:allmethods}, volume-based methods can handle sudden depth changes, but the reconstruction results are noisy, suffering more from the shape-radiance ambiguity. This phenomenon is more evident in textured regions. For instance, in the third row of \cref{fig:allmethods}, the leather side of \textit{Rolleiflex} attains a much rougher surface than the metal frame. In contrast, with the inductive bias of SDF, NeuS tends to learn a smoother surface to compensate for the effects of dynamic illuminations, thus lacking geometric details.

Our experiments show that the geometry of image-based reconstruction is affected by several factors: inconsistent lighting, lack of texture, reflections, etc. However, with view-dependence normalization, our method suffers less from these problems and outperforms by a large margin. With the employment of $\mathcal{F}$ in the architecture, both NeuS and VolSDF achieve around 10\% improvement in IoU and f-score, and up to 30\% improvement in Chamfer $\mathcal{L}1$ Distance, demonstrating the generality of our methods in enhancing geometry with different reconstruction pipelines.

\textbf{Robustness to lighting condition.} Quantitative evaluations in \cref{tab:vary-in-lights} show that our method is more consistent in surface quality when the illumination changes. As the lighting condition becomes unstable, NeuS \cite{wang2021neus} drops a lot in geometry quality measured by IoU scores, whereas our method only drops by a graceful margin. Moreover, with the proposed view-dependence normalization, our method can further improve in the static lighting scenario.

\textbf{Novel view rendering.} \cref{fig:rolleiflex_rendered} shows novel-view image generated by NeuS \cite{wang2021neus} and our method. As our method biases the SDF network towards enhanced geometry, better appearance reconstruction is achieved simultaneously.

\begin{figure}
    \centering
    \includegraphics[width=.96\linewidth]{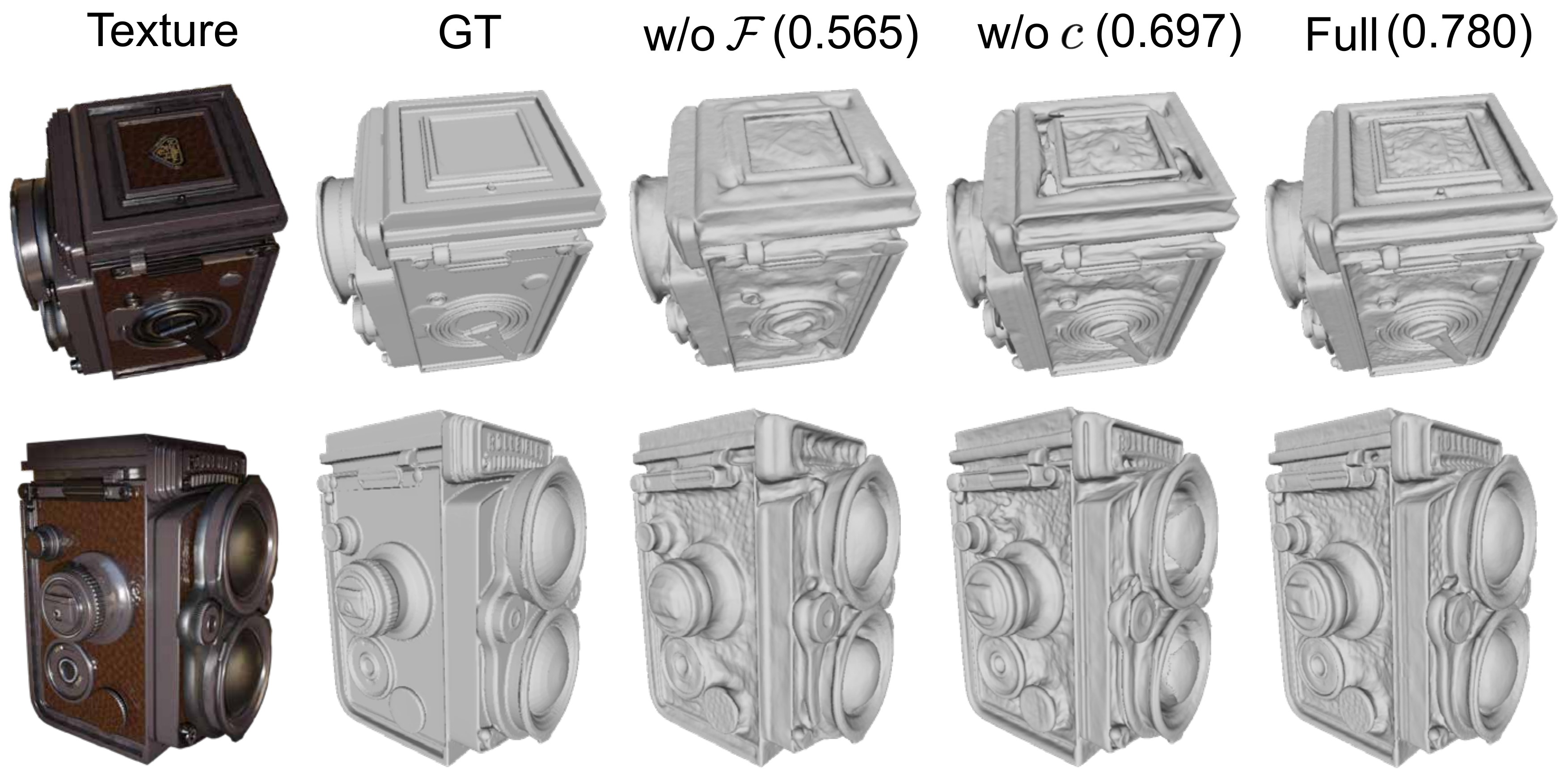}
    \vspace{-0.3cm}
    \caption{Effectiveness of the feature function $\mathcal{F}$ and the color function $c$. F-scores are shown in the brackets. Lacking the proposed view-dependence normalization incurs a larger loss in geometric quality and results in over-smoothed surfaces.}
    \vspace{-0.2cm}
    \label{fig:ablation-cnd}
\end{figure}

% \begin{figure}
%   \begin{subfigure}{1.0\linewidth}
%     \centering
%     \includegraphics[width=\linewidth]{figures/linglan-scale.pdf}
%     \caption{Ablation on resolution of features $d_k^r$. As the scale of changes from full ($-f$) to half ($-h$), object with sharp edges shows variation as 'dilation'.}
%     \label{fig:linglan}
%   \end{subfigure}
    % \caption{Ablation studies.}

\begin{table}[]
    \centering
    \resizebox{.8\linewidth}{!}
    {
    \begin{tabular}{@{}cccccc@{}}
        \toprule
         & IoU $\uparrow$ & L1 CD $\downarrow$ & L2 CD $\downarrow$  & NC $\uparrow$ & f-score $\uparrow$ \\
        \midrule
        w/o & 0.640 & 0.558 & 1.466 & 0.834 & 0.763 \\
        $l=1$ & \textbf{0.705} & \textbf{0.396} & \textbf{0.823} & \textbf{0.858} & \textbf{0.851} \\
        $l=2$ & 0.687 & 0.434 & 1.032 & 0.854 & 0.830 \\
        $l=3$ & 0.682 & 0.419 & 0.842 & 0.854 & 0.827 \\
        $l=4$ & 0.601 & 0.536 & 1.038 & 0.834 & 0.717 \\
        \bottomrule
    \end{tabular}
    }
    \vspace{-0.2cm}
    \caption{Effect of using depth features from different encoder layers ($l$) of the distillation network $\psi$.}
    \vspace{-0.4cm}
    \label{tab:ablation-on-l}
\end{table}

\subsection{Ablations}
% we compare our model with three variants on $Rolleiflex$: Model trained without $\mathcal{F}$; Model trained with both color branch and feature function $\mathcal{F}$ (ModelB); and Model trained with feature function $\mathcal{F}$ (ModelC).
% We report qualitative result and corresponding f-score in \cref{fig:ablation-cnd}. ModelB outperforms ModelA and ModelC, the latter ones faces a high probability of failure, where the geometry shows artifacts. Combination of feature branch $\mathcal{F}$ and color branch partly reduces the demand for well-trained depth estimation network, and offers the best geometry.

{\bf Color vs. Feature}: \cref{fig:ablation-cnd} shows that without the normalization feature function that regularizes the SDF network (w/o $\mathcal{F}$), the reconstruction drops substantially, showing the benefits of our directional view-dependence normalization technique. Qualitatively, the feature field enables capturing sharp geometry. Removing the color function (w/o $c$) also produces worse geometries. Thus, the radiance field can help rectify abnormalities in case of misleading depth features. Moreover, with the combination of $F$ and $c$, our full model achieves the best performance.

In \cref{tab:ablation-on-l}, we perform an ablation study on how the performance changes depending on where the depth feature $\psi^l$ is extracted. Restricted by the GPU memory, we only experiment with the first four layers from the encoder of $\Psi$. The features from each layer are in dimensions 96, 96, 192, and 384, with sizes of 1/2, 1/4, 1/8, and 1/16 of the input. We upsample input images to ensure the features are of the same resolution. Results of the model trained without $\psi^l$ are reported in the first row (w/o).
Generally speaking, models trained with smaller $l$ gain better from the view-dependence normalization. Notably, the quality decreases significantly when $l$ increases to 4,
which indicates that high-level depth features may be too invariant to recover the underlying structures.

\begin{figure}
  \begin{subfigure}{1.0\linewidth}
    \centering
    \includegraphics[width=\linewidth]{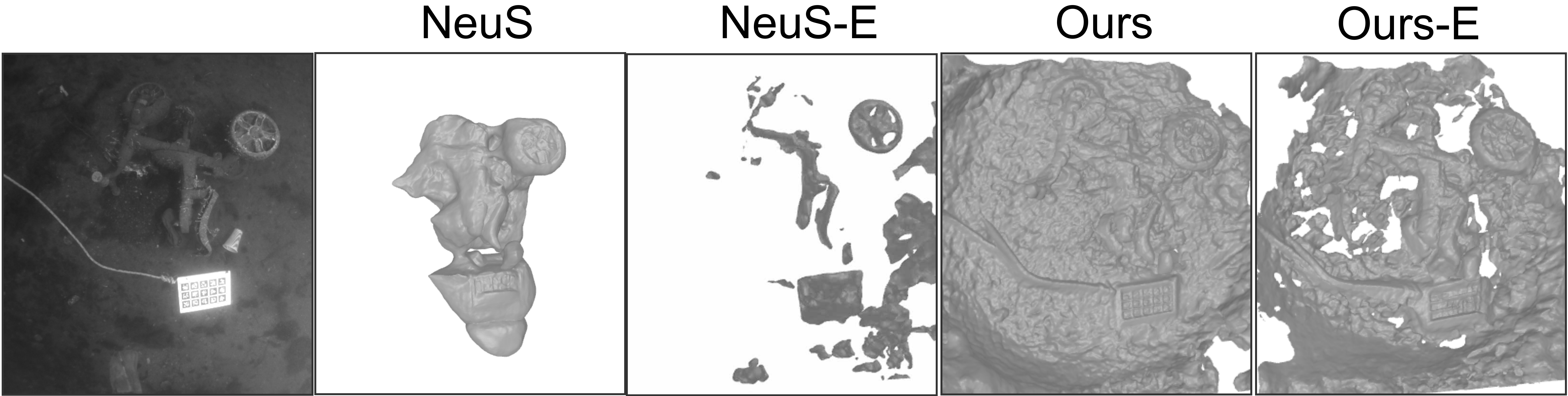}
    \caption{Reconstructed scenes from the underwater dataset AQUALOC \cite{ferrera2019aqualoc}. \textit{E} stands for using image enhancement to improve the contrast. Despite the enhancement, NeuS \cite{wang2021neus} still overlooks geometry details. Our method can better deal with the low-illuminance scenario.
    }
    \label{fig:aqualoc-0}
  \end{subfigure}
  \vspace{0.1cm}
  \begin{subfigure}{1.0\linewidth}
    \centering
    \includegraphics[width=\linewidth]{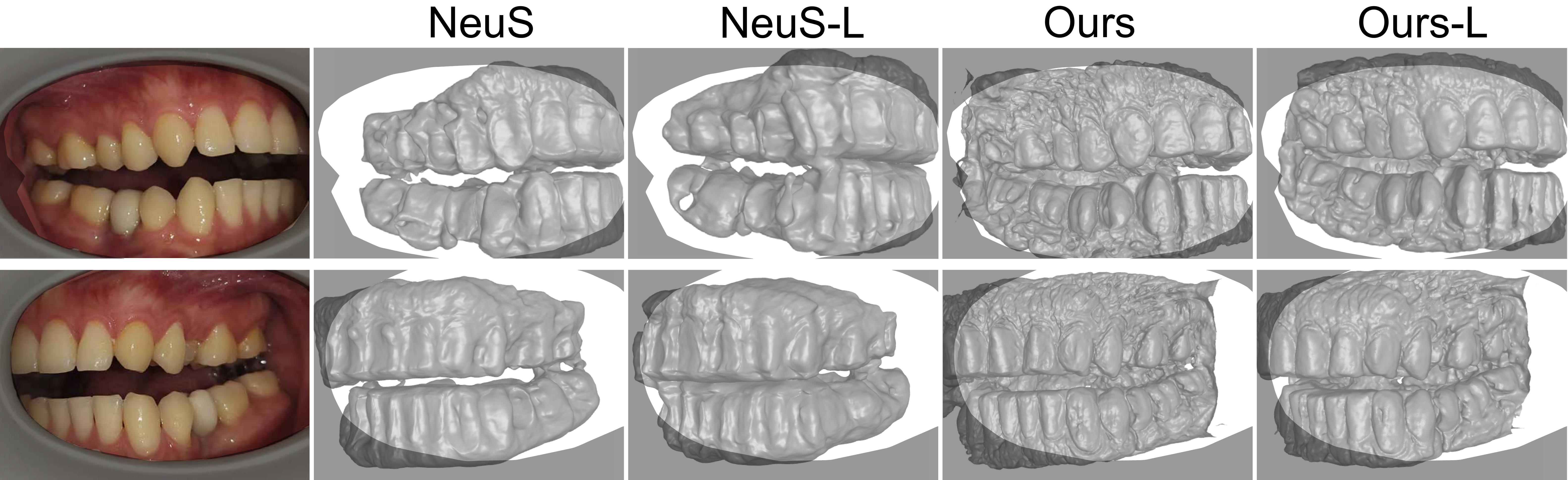}
    \caption{Reconstructed geometry from intra-oral images. \textit{L} means training with camera pose optimization. Our method performs well in such complex scenes, while NeuS \cite{wang2021neus} fails to reason about the teeth structure.}
    \label{fig:skd2}
  \end{subfigure}
    \vspace{-0.4cm}
    \caption{Results on unconstrained real-world scenes with dynamic lighting show the practical usefulness of our method for dim-environment applications.}
    \vspace{-0.4cm}
    \label{fig:realdata}
\end{figure}

\subsection{Real-world capture}
\vspace{-0.1cm}
% teeth_sccan/skd2
% depth network trained on synthetic teeth data
% aqualoc
% To test the capability of decoupling of shape-radiance ambiguity and directional view-dependence in practical application, we also conduct experiments on two real dataset, one intra-oral scanning data and another underwater dataset AQUALOC \cite{ferrera2019aqualoc}.

% images selected from the harbor video sequences (at a depth of a few meters) of \cite{ferrera2019aqualoc}, the data is recorded by Remotely Operated Vehicles equipped with a monocular monochromatic camera, a MEMS-IMU and a pressure sensor, embedded in a single enclosure.

% We perform experiment on real data 
% and results are reported in \cref{fig:realdata}. 
% %The two
% Results are reconstructions from underwater video sequences and intra-oral scans respectively. 
Geometry reconstruction results from the underwater sequences in 
% \cref{fig:aqualoc-0}
Fig.~\ref{fig:realdata} (a) show that our model can have robust geometry estimation under harsh conditions, for instance, dark scenes with a weak moving light source.
For NeuS, contrast enhancement on input images (NeuS-E) could help to learn better geometry, but the quality still lags. In contrast, our method can generate a reasonable surface without further modification of the pipeline.

% These results shows that our model is promising to apply in submarine scenarios.

Fig.~\ref{fig:realdata} (b) shows another difficult scenario, i.e., intra-oral scans. To achieve better reconstruction quality, we further train the models with camera pose optimization.  The occlusion by mouth gag, dynamic lighting, and the lack of structural information incur challenges for NeuS. As expected, our model learns a more accurate geometry for teeth despite textureless and highly reflective surfaces.

% \begin{figure}
%   \begin{subfigure}{1.0\linewidth}
%     \centering
%     \includegraphics[width=0.8\linewidth]{figures/AQUALOC.png}
%     \caption{Real data1. images selected from the harbor video sequences (at a depth of a few meters) of \cite{ferrera2019aqualoc}, the data is recorded by Remotely Operated Vehicles equipped with a monocular monochromatic camera, a MEMS-IMU and a pressure sensor, embedded in a single enclosure. }
%     \label{fig:aqualoc5}
%   \end{subfigure}
%     \caption{Caption}
% \end{figure}

\vspace{-0.1cm}
\section{Discussion}
\vspace{-0.1cm}

We show that the coupling of the view-dependent radiance and the shape-radiance ambiguity -- causing issues when quality geometry is important -- can be alleviated by a simple yet effective view-dependence normalization. 
The effectiveness is guaranteed by enforcing a single optimal directional capacity that maximizes the rendering before the shape-radiance ambiguity kicks in. 
The view-dependence normalization comes down to distilling robust and discriminative information from the radiance field without relying on external supervision. 
We demonstrated its superiority in challenging scenes, e.g., with reflective and textureless surfaces under dynamic lighting conditions. 
We hope the proposed technique can serve future research in applying neural radiance fields for (unconstrained) real-world geometric reasoning tasks.

\vspace{0.3cm}
\noindent {\bf Acknowledgement}: We thank the support of a grant from the Stanford Human-Centered AI Institute, an HKU-100 research award, and a Vannevar Bush Faculty Fellowship.

\newpage
%%%%%%%%% REFERENCES
{\small
\normalem
\bibliographystyle{cvpr23/ieee_fullname}
\bibliography{ref}
}

\end{document}